\documentclass[letterpaper, 10 pt, conference]{ieeeconf}  

\IEEEoverridecommandlockouts

\IEEEoverridecommandlockouts                              

\usepackage{graphics}
\usepackage{amsmath}
\usepackage{amssymb}
\usepackage[colorlinks, linkcolor=blue, citecolor=blue]{hyperref}
\usepackage{cite}
\usepackage{algorithm}
\usepackage{url}
\usepackage{algpseudocode}
\usepackage{epsfig}
\usepackage{graphicx}
\usepackage{caption}
\usepackage{subfigure}
\usepackage{mathrsfs}
\usepackage{array}
\usepackage{multirow}
\usepackage{multicol}
\usepackage{booktabs}
\usepackage{makecell}
\usepackage{xcolor}
\usepackage{color}
\usepackage{amsfonts}
\usepackage{gensymb}
\usepackage{threeparttable}
\usepackage{textcomp}

\usepackage{eso-pic}

\IEEEaftertitletext{\vspace{-1em}}

\title{\LARGE \bf
Revisiting Topological Graphs for Macro Action based\\Closed-loop Reinforcement Learning of Vision Language Navigation in\\Continuous Environment
}

\author{
Shuhao Ye$^{1}$, Sitong Mao$^{2}$, Yuxiang Cui$^{3}$, Yufei Wei$^{1}$, Xuan Yu$^{1}$, Shichao Zhai$^{1}$, Wen Chen$^{2}$, Shunbo Zhou$^{2}$, \\
Rong Xiong$^{1}$, Yue Wang$^{1}$$^{\dagger}$
\thanks{This work was supported by the Fundamental and Interdisciplinary Disciplines Breakthrough Plan of the Ministry of Education of China (Grant No. JYB2025XDXM208), the National Natural Science Foundation of China under Grants 62522317 and 62373322, the Key R\&D Program of Hangzhou under Grant No. 2024SZD1A30, and the Zhejiang Provincial Natural Science Foundation of China under Grant No. LD25F030001.}%
\thanks{$^{1}$Zhejiang University, Hangzhou, China}%
\thanks{$^{2}$Huawei Technologies Co., Ltd}%
\thanks{$^{3}$Zhejiang Humanoid Robot Innovation Center, Ningbo, China}%
\thanks{$\dag$Corresponding author: Yue Wang (email: wangyue@iipc.zju.edu.cn)
}
}

\begin{document}

\maketitle

\thispagestyle{empty}
\pagestyle{empty}

\AddToShipoutPictureFG*{%
  \AtPageLowerLeft{%
    \put(
      \LenToUnit{\dimexpr1in+\hoffset+\oddsidemargin\relax},
      \LenToUnit{6mm}
    ){%
      \raisebox{\depth}{%
        \begin{minipage}[b]{\textwidth}
          \normalfont\fontsize{8}{9.5}\selectfont
          \color{black}\raggedright
          \setlength{\parindent}{0pt}%
          \textcopyright~2026 IEEE. Personal use of this material
          is permitted. Permission from IEEE must be obtained for
          all other uses, in any current or future media, including
          reprinting/republishing this material for advertising or
          promotional purposes, creating new collective works,
          for resale or redistribution to servers or lists, or
          reuse of any copyrighted component of this work in
          other works.
          \par
        \end{minipage}%
      }%
    }%
  }%
}

\begin{abstract}

Vision-Language Navigation in Continuous Environments (VLN-CE) requires an agent to follow natural language instructions through unseen environments. Existing imitation learning (IL) pipelines struggle in this closed-loop setting: behavior cloning suffers from distribution shift, and DAgger's expert actions become ambiguous upon trajectory deviation. While Reinforcement Learning (RL) offers a natural paradigm to address this, directly applying RL to micro action spaces is sample-inefficient due to reward sparsity. To overcome this bottleneck, we reformulate VLN-CE as a Hierarchical Markov Decision Process (MDP), explicitly decoupling high-level planning from low-level control. By abstracting the environment into a topological graph, our high-level policy operates on a macro action space of frontier nodes, with a training-free low-level controller acting as its state transition, which significantly compresses the decision horizon and makes closed-loop RL tractable. To support RL optimization on the macro MDP, we propose an action-aware value head to effectively evaluate state values under the dynamic frontier action space, powering a graph-based PPO. Extensive experiments demonstrate the effectiveness of our architecture. Finally, our model achieves state-of-the-art performance on the R2R-CE and RxR-CE benchmarks.
\end{abstract}

\section{Introduction}

Vision-Language Navigation in Continuous Environments (VLN-CE) \cite{krantz2020beyond} tasks an agent with following natural language instructions through unseen environments, executing hundreds of micro actions. This long‑horizon closed‑loop process suffers from a fundamental challenge: distribution shift between training and test conditions leads to error accumulation and eventual task failure.

Existing approaches tackle distribution shift from two perspectives. Graph‑based methods \cite{an2024etpnav, an2022bevbert} abstract the environment into topological nodes and define macro actions (selecting frontiers), compressing the decision horizon to merely 5–20 choices. This structural abstraction truncates error accumulation at its source. Large Vision‑Language Models (LVLMs) \cite{zhang2024uninavid, wei2025streamvln} instead rely on scaling: internet‑scale pretraining provides strong semantic priors that help generalize to shifted distribution, even while operating on the original micro action space, making it a more popular trend.

\begin{figure}[t]
    \centering
    \begin{minipage}[t]{1\linewidth}
    \centering
    \includegraphics[width=1\linewidth]{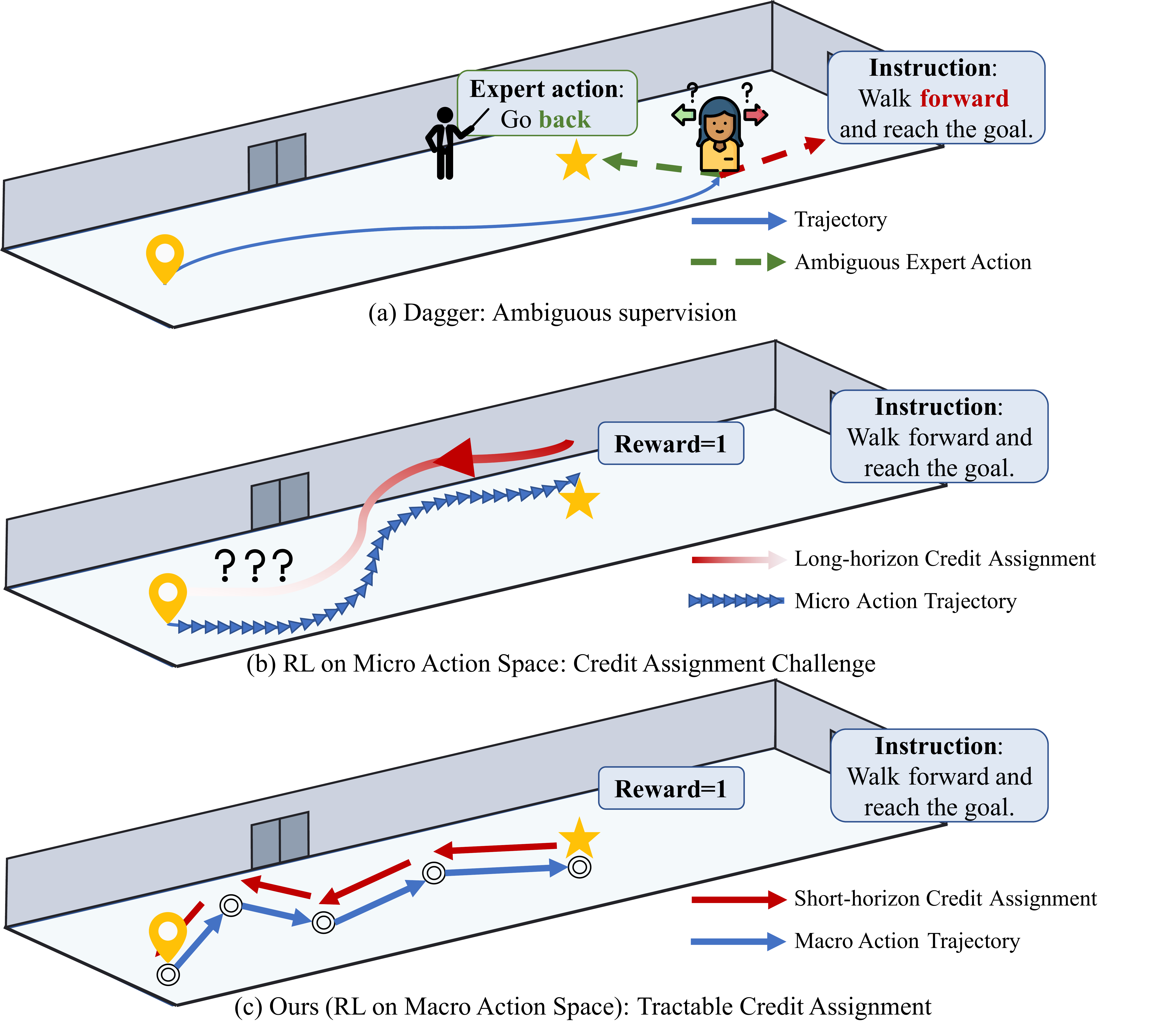}
    \end{minipage}
    \caption{\small{Comparison of three training paradigms in VLN-CE. DAgger's expert actions for error recovery may conflict with instructions, and closed-loop RL in micro action spaces faces credit assignment challenges. We propose closed-loop RL in macro action spaces to achieve tractable credit assignment.}}
    \label{fig:intro}
    \vspace{-20pt}
\end{figure}

In addition to architectural designs, to further relieve distribution shift, DAgger \cite{dagger} is usually added after behavior cloning (BC) in the learning process for both lines of methods. However, DAgger assumes access to an oracle expert that can provide correct actions in arbitrary states. This assumption breaks down in VLN-CE because optimal expert actions become ambiguous the moment an agent deviates from the reference trajectory, suggesting that IL is mismatched for closed‑loop navigation. Reinforcement learning (RL) offers a more natural paradigm by learning directly from task outcomes, but applying RL directly to LVLMs is sample-inefficient. The main reason is the long horizon due to the micro action space. To circumvent this, prior RL attempts generally rely on restrictive workarounds: either constraining optimization to single-turn language output with offline trajectory datasets (open-loop RL) \cite{qi2025vlnr1} or employing distance heuristics as dense reward shaping \cite{gao2025octonav}. By substituting pure environmental trial-and-error with rigid dataset-alignment or greedy distance minimization, these compromises bypass autonomous exploration, preventing RL from realizing its full potential.

This dilemma raises a question:  \textit{Can we leverage a macro action space to improve sample efficiency and apply the closed-loop RL?} To answer this, we revisit the structural abstraction of topological graphs, reformulating VLN-CE as a hierarchical decision-making problem modeled via a hierarchical Markov Decision Process (MDP). This decouples high-level planning from low-level control. A high-level policy selects macro actions over frontiers, reducing the decision horizon to make closed-loop RL highly sample-efficient. Meanwhile, the low-level controller becomes a short-horizon routing task handled by off-the-shelf methods, serving as the environmental state transition for the macro MDP. Ultimately, this formulation transforms the RL problem into a tractable one.


Building upon this insight, we instantiate our high-level planner with a Graph Transformer policy, leaving the low-level controller to a heuristic rotate-then-forward strategy. 
To support RL optimization on the macro MDP, we propose an action-aware value head to effectively evaluate state values under the dynamic frontier action space, powering a graph-based PPO.
Finally, following pretraining and DAgger initialization, we integrate RL into the high-level policy by adopting a closed-loop Reinforcement Fine-Tuning (RFT) paradigm.
Extensive experiments evaluate various RL algorithms within our architecture, identifying Proximal Policy Optimization (PPO) \cite{PPO} as the most effective. Our framework outperforms baselines trained on micro action space. Furthermore, we show that RFT and dataset scaling provide orthogonal performance gains. 

Our main contributions are summarized as follows:
\begin{itemize}
\item We reformulate VLN-CE as a hierarchical decision-making problem modeled by hierarchical MDP and propose to apply RL to the macro MDP.
\item We propose an efficient closed-loop RFT framework on the macro action space of topological frontiers.
\item We devise a graph-based PPO, featuring an action-aware value head to effectively evaluate state values under dynamic action space inherent to topological graphs.
\item Extensive experiments demonstrate the effectiveness of our architecture. Our model reaches SOTA performance on the R2R-CE and RxR-CE datasets.
\end{itemize}

\section{Related Work}

\subsection{VLN in Continuous Environments (VLN-CE)}

To address VLN-CE task \cite{krantz2020beyond}, early approaches attempted to directly map visual and textual inputs to micro actions \cite{law}, but their performance was often limited by inefficient action space. The introduction of the waypoint predictor \cite{krantz2021waypoint} marked a significant evolution, shifting the problem from low-level control to high-level waypoint selection. This waypoint-based paradigm was further formalized by \cite{an2024etpnav}, which organize waypoints into a topological graph and employ a Transformer-based planner for high-level planning. While subsequent works within this framework \cite{an2022bevbert, hnr} have focused on improving visual representations to aid planning, the performance gains from stacking more powerful feature extractors have begun to plateau. 

In parallel, the rapid advancement of Large Language Models has inspired a new line of research using LLMs/LVLMs as navigation agents \cite{wei2025streamvln, qi2025vlnr1}. The powerful pretrained weights of these models afford a deeper understanding of language instructions. However, current LVLM-based methods are typically limited in two ways: they have difficulty processing panoramic images, often relying on a monocular view, and they cannot effectively leverage topological structures, resorting to the history of image frames as their environmental representation.


\subsection{Reinforcement learning (RL) in VLN-CE}

While RL was easily applied in discrete VLN \cite{chen2022reinforced}, its direct application to VLN-CE is expensive due to the micro action space and longer trajectory length. Krantz et al. \cite{krantz2021waypoint} first apply RL to the waypoint predictor. Although this reduced the learning complexity, the agent still faced the difficult dual task of generating waypoints that were both navigable and instruction-compliant. Recently, the graph-based VLN \cite{an2024etpnav} further decouples waypoint generation from instruction following, which is an ideal framework for implementing RL.

Concurrently, the training paradigms of LVLMs \cite{shao2024deepseekmath} have shown that SFT followed by RL—a process known as RFT—is a highly efficient approach. Qi et al. \cite{qi2025vlnr1} introduced RFT to a LVLM-based policy. However, as their approach is based on offline training, the RFT is limited to fine-tuning a single-turn language output with the offline trajectory datasets. This renders their RFT process actually open-loop with respect to the navigation task. Gao et al. \cite{gao2025octonav} recognized the above problem and attempted to directly perform closed-loop RFT on LVLMs based on micro action space. However, the decision horizon of hundreds of steps necessitates the use of distance-to-goal-based dense rewards to address the sparse reward problem, which still leads to a mismatch between trajectories and instructions, thereby partially undermining the significance of RL.
In contrast, our work introduces a closed-loop RFT paradigm based on a macro action space, which obviates the need for dense reward shaping.

\begin{figure*}[h]
    \centering
    \begin{minipage}[t]{1\linewidth}
    \centering
    \includegraphics[width=1\linewidth]{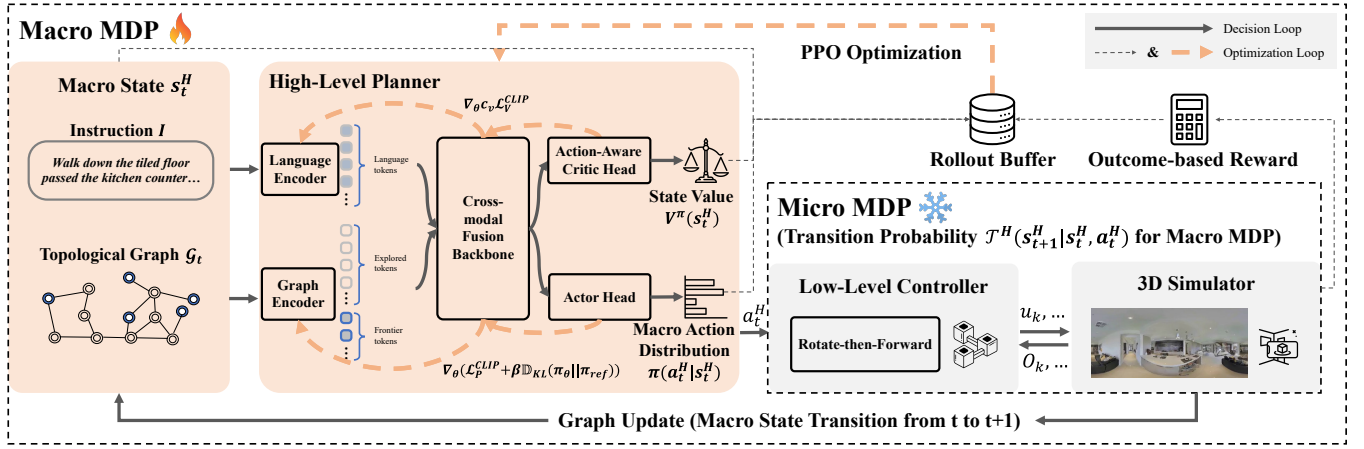}
    \end{minipage}
    \caption{\small{Overview of our framework. We reformulate the VLN-CE task into a hierarchical MDP based on the topological graph, perform closed-loop RL on the Macro MDP, and use the Micro MDP as its state transition probability.}}
    \label{fig:overview}
    \vspace{-15pt}
\end{figure*}

\section{Method}

\subsection{Problem Formulation: A Hierarchical MDP Perspective}
In VLN-CE, an agent navigates towards a target specified by a natural language instruction $I$. At each timestep $t$, the agent receives a panoramic RGBD observation $O_t$ (comprising 12 egocentric views) and output a base motor command $u_t$ from a discrete micro action space $\mathcal{U}$ (\textit{move forward, turn left, turn right, stop}).

Formulating this entire process as a single MDP over $\mathcal{U}$ yields decision horizons exceeding hundreds of steps, which leading to reward sparsity for RL. To make RL tractable, we reformulate VLN-CE as a Hierarchical MDP that decouples the problem into high-level planning and low-level control. We focus our RL optimization on the high-level policy, while leaving the low-level execution to a heuristic controller. As shown in Figure \ref{fig:overview}.

\textbf{High-Level Planner (Macro MDP):}
The upper layer abstracts the long-horizon task using a topological graph, compressing the decision sequence from hundreds of steps to merely 5-20 steps. We define this macro MDP as a tuple $\langle \mathcal{S}^H, \mathcal{A}^H, \mathcal{T}^H, \mathcal{R}^H, \gamma^H \rangle$:
\begin{itemize}
\item \textbf{State ($\mathcal{S}^H$):} The macro state $s^H_t = (I, \mathcal{H}_t) \in \mathcal{S}^H$ consists of the language instruction $I$ and the graph trajectory $\mathcal{H}_t$, which is incrementally constructed from panoramic observations up to high-level step $t$.
\item \textbf{Action ($\mathcal{A}^H$):} The macro action $a^H_t \in \mathcal{A}^H_t$ is defined as selecting a subgoal from the dynamic set of unvisited frontier nodes, alongside a special STOP action.
\item \textbf{Transition ($\mathcal{T}^H$):} The transition $\mathcal{T}^H(s^H_{t+1} | s^H_t, a^H_t)$ is implicitly governed by the low-level controller. It updates the graph by incorporating new observations once the agent reaches the frontier $a^H_t$.
\item \textbf{Reward ($\mathcal{R}^H$) and Discount ($\gamma^H$):} The agent receives an outcome-based reward $r(s^H_T)$ at the end of the episode evaluating the final trajectory's alignment with the instruction. $\gamma^H \in (0, 1]$ is the discount factor.
\end{itemize}

\textbf{Low-Level Controller (Micro MDP):}
Once a high-level frontier $a^H_t$ is selected, the lower layer initiates a short-horizon routing task. We formulate this local execution as a micro MDP $\langle \mathcal{S}^L, \mathcal{A}^L, \mathcal{T}^L, \mathcal{R}^L, \gamma^L \rangle$:
\begin{itemize}
\item \textbf{State ($\mathcal{S}^L$):} The micro state $s^L_k$ comprises the current instantaneous panoramic observation $O_k$ and the agent's relative pose to the assigned macro subgoal $a^H_t$.
\item \textbf{Action ($\mathcal{A}^L$):} The micro action space $\mathcal{A}^L \equiv \mathcal{U}$.
\item \textbf{Transition ($\mathcal{T}^L$):} The transition $\mathcal{T}^L(s^L_{k+1} | s^L_k, u_k)$ is determined by the simulator or physical environments.
\item \textbf{Reward ($\mathcal{R}^L$) and Discount ($\gamma^L$):} $\mathcal{R}^L$ is defined as a dense reward indicating progress towards the subgoal $a^H_t$, and $\gamma^L$ is the low-level discount factor.
\end{itemize}
While this micro MDP could be optimized via RL, existing methods can also solve this problem. Thus, we directly employ a deterministic heuristic policy for simplicity, avoiding low-level RL training.

\subsection{Closed-loop RFT via Graph-based PPO}
\label{sec:rft}

Having isolated the RL optimization to the macro MDP, we now introduce the specific algorithm used to finetune the IL-initialized planner. While previous RFT efforts in vision-language tasks employ Group Relative Policy Optimization (GRPO) \cite{shao2024deepseekmath}, it only assigns a single trajectory-level advantage, failing to provide the step-level credit assignment. Conversely, off-policy algorithms (e.g., SAC \cite{haarnoja2018sac}) suffer from outdated data in replay buffers, which destabilizes the delicate alignment of cross-modal representations. Therefore, we adopt Proximal Policy Optimization (PPO) \cite{PPO}, an on-policy algorithm that enables fine-grained credit assignment via step-wise state value estimation $V(s)$.

\subsubsection{Action-Aware Value Estimation}
Estimating the state value in our graph-based macro MDP introduces a structural challenge. According to the Bellman equation, the state value $V^\pi(s^H_t)$ is the expected return over the available action space: $V^\pi(s^H_t) = \sum_{a \in \mathcal{A}^H_t} \pi(a|s^H_t) Q^\pi(s^H_t, a)$.

Unlike standard MDPs with a static action set, our macro action space $\mathcal{A}^H_t$ (the generated unvisited frontiers) is dynamic. If the agent reaches a topological bottleneck where all frontiers in $\mathcal{A}^H_t$ lead to dead ends, the action values $Q^\pi$ for all current candidates will be inevitably low, causing $V^\pi(s^H_t)$ to plummet. Consequently, a traditional Critic network that predicts $V$ based solely on the historical state $s_t^H = (I, \mathcal{H}_t)$ without observing the dynamic action space suffers from partial observability. It cannot distinguish a well-connected hub from a topological dead-end until the next step.

To resolve this, we propose an Action-Aware Value Estimation architecture. We explicitly incorporate the dynamic action space $\mathcal{A}^H_t$ as an intrinsic, forward-looking feature of the state. Specifically, we expand the graph trajectory $\mathcal{H}_t$ into a holistic topological graph $\mathcal{G}_t$ that encapsulates both the explored history and the unvisited frontiers $\mathcal{A}_t^H$. Correspondingly, we augment the high-level state definition to $s_t^H = (I, \mathcal{G}_t)$. By evaluating this action-aware state, the Critic can predict the state value $V^\pi(s_t^H)$ conditioned on the currently available future paths.


Additionally, since the policy network (Actor) already requires the explicit features of $\mathcal{A}^H_t$ to compute action probabilities over the frontiers, both the Actor and the Critic can share the same cross-modal representation backbone. We simply append a lightweight, scalar Value Head parallel to the policy head. This shared-backbone design resolves the partial observability of the Critic while introducing negligible computational and memory overhead during the RFT stage.

\subsubsection{Outcome-Driven Reward and GAE}
To optimize this architecture, we design a outcome-driven terminal reward $R_T$ to evaluate the entire trajectory:
\begin{equation}
\label{eq:reward}
R_T = \text{Success} + \text{SPL} + \text{nDTW}
\end{equation}
Here, the binary $\text{Success}$ indicator encourages the agent to halt within 1.5m of the target. $\text{SPL}$ (Success weighted by Path Length) \cite{anderson2018evaluation} penalizes detours, promoting concise navigation. While $\text{nDTW}$ (normalized Dynamic Time Warping) \cite{ilharco2019general} measures the fidelity between the agent's and reference path. Even in failed episodes where $\text{Success}$ and $\text{SPL}$ are zero, $\text{nDTW}$ provides a positive reward signal that effectively distinguishes near-misses from completely divergent trajectories, guiding the early stages of RL exploration.

With the defined reward function, we compute the Temporal Difference (TD) error $\delta_t$ as follows:
\begin{equation}
\delta_t =
\begin{cases}
\gamma^H V^\pi(\mathcal{G}_{t+1}) - V^\pi(\mathcal{G}_t), & \text{if } t < T \\
R_T - V^\pi(\mathcal{G}_T), & \text{if } t = T \\
\end{cases}
\end{equation}
Using these TD errors, we then employ Generalized Advantage Estimation (GAE) to calculate the advantage $\hat{A}_t = \sum_{l=0}^{T-t} (\gamma^H \lambda)^l \delta_{t+l}$. Finally, this advantage is utilized to update the shared-backbone network via the PPO objective.


\subsection{Hierarchical Policy Architecture}
To instantiate the hierarchical MDP formulated in the previous section, our system comprises three modules: an online graph constructor that builds the topological graph $\mathcal{G}_t$, a multimodal high-level planner that evaluates this state to output macro action logits and a scalar value, and a heuristic low-level controller for execution.

\subsubsection{Topological Graph Construction}
The topological graph $\mathcal{G}_t = (\mathcal{V}_t, \mathcal{E}_t)$ serves as the agent's spatial memory and action space. The node set $\mathcal{V}_t = \mathcal{H}_t \cup \mathcal{A}_t^H$ consists of explored nodes $\mathcal{H}_t$ (the agent's starting position and all previous macro action endpoints) and unvisited frontier nodes $\mathcal{A}_t^H$. At each explored node, a waypoint predictor \cite{hong2022bridging} extracts surrounding navigable locations as frontiers, pruning those excessively close to existing nodes. Each explored node stores 12 panoramic RGB-D images, while each frontier node is bound to a specific angular grid corresponding to the visual observation from its parent explored node.

\subsubsection{Modality Encoding}
The language instruction $\mathcal{I}$ is tokenized and processed by a standard BERT-style encoder into the sequence of text features $I$.

For the topological graph $\mathcal{G}_t$, the 12 RGBD views of each explored node are processed via a visual backbone and a panorama transformer. To mitigate the computational overhead, we compress these 12 view features into four directional tokens (front, back, left, right) using attention pooling with learnable queries.

The feature for a frontier node is directly inherited from the specific angular grid feature of its parent explored node. Consequently, each frontier is encoded as one token vector, establishing a strict one-to-one mapping between the frontier tokens and the macro action space $\mathcal{A}_t^H$. Finally, we inject a step embedding to encode chronological order and a pose embedding to capture the relative spatial pose to the current node, forming the initial graph feature sequence $G$.

\subsubsection{Graph-Aware Cross-Modal Fusion}
The encoded features $I$ and $G$ are fed into a Cross-Modal Fusion Network. Each layer sequentially performs cross-attention between modalities and self-attention within modalities.

To explicitly capture the spatial relationships between nodes, we replace the standard self-attention for graph features with Graph-Aware Self-Attention (GASA) \cite{an2024etpnav}. We construct a spatial bias matrix $M$ based on the topological path distance $d_{ij}$ and the relative orientation $\Delta\theta_{ij}$ between token $i$ and $j$. To ensure angular continuity, the orientation is mapped to a 2D vector $[\sin(\Delta\theta_{ij}), \cos(\Delta\theta_{ij})]$, concatenated with distance to form $v_{ij} = [d_{ij}, \sin(\Delta\theta_{ij}), \cos(\Delta\theta_{ij})]^T$, and processed via an MLP:
\begin{equation}
    M_{i,j} = W_2 \cdot \text{GELU}(W_1 \cdot v_{ij} + b_1) + b_2
\end{equation}
This structural bias $M$ is directly added to the attention logits following GASA.

Following the cross-modal fusion (yielding intermediate features $I_{\text{cross}}$ and $G_{\text{cross}}$), we distill navigational cues from the text to enhance the graph representation:
\begin{equation}
    G_{\text{guide}} = \text{CrossAttn}(G_{\text{cross}}, I_{\text{cross}})
\end{equation}

The final topological representation is obtained by:
\begin{equation}
    G_{\text{out}} = \text{Concat}\left(G_{\text{cross}}, \text{FFN}(G_{\text{guide}})\right)
\end{equation}

\subsubsection{Dual-Head Decision and Low-Level Control}
The refined representation $G_{\text{out}}$ supports the dual-head decision process. The actor head uses a FFN to score only the specific tokens corresponding to the frontier nodes (and the STOP token), yielding the macro action logits $s_{i}$:
\begin{equation}
    s_{i} = \text{FFN}_{\text{actor}}(g_{i})
\end{equation}

Concurrently, to realize the Action-Aware Value Estimation formulated in Section \ref{sec:rft}, a Value Head computes the scalar state value $V^\pi(s^H_t)$. We introduce a single learnable query $q_v$ that performs attention pooling over the entirety of $G_{\text{out}}$. By aggregating the explored history and the frontier tokens, this operation compresses the topological graph into a single vector, which is then projected into the value scalar:
\begin{gather}
    V^\pi(s^H_t) = \text{FFN}_{\text{value}}(\text{CrossAttn}(q_{v}, G_{\text{out}}))
\end{gather}

Once a target frontier is selected by the policy, the low-level controller executes a rotate-then-forward with tryout strategy \cite{an2024etpnav}. If the selected subgoal belongs to a distant explored node, the controller first plans a topological path, sequentially navigating node-by-node until the terminal frontier is reached or the low-level controller quits due to collisions that can't be resolved by the tryout strategy. After that, a new explored node is instantiated and surrounding frontiers are updated regardless of execution success.

\subsection{Training Paradigm}
\label{sec:train}
We train our high-level planner through a progressive three-stage curriculum.

Stage 1 \& 2: Pretraining and DAgger with Value Warm-up. To establish initial visual-linguistic alignment and navigational priors, we adopt the pretraining and DAgger setups as in \cite{an2024etpnav}, with one key addition: we simultaneously warm up our Action-Aware Value Head during the DAgger phase. The joint loss in DAgger stage is defined as:
\begin{equation}
    \mathcal{L}_{\text{DAgger}} = \mathcal{L}_{\text{CE}} + c_v \mathcal{L}_{V}^{\text{CLIP}}
\end{equation}
where $\mathcal{L}_{\text{CE}}$ is the cross-entropy loss for imitation, $\mathcal{L}_{V}^{\text{CLIP}}$ is the standard clipped MSE loss, and $c_v$ is the value coefficient.

Stage 3: Closed-Loop RFT. To further refine planning capabilities, we transition to the PPO optimization detailed in Section \ref{sec:rft}. To prevent policy degradation, we regularize the update with a KL divergence loss against the DAgger-trained reference model $\pi_{\text{ref}}$. The shared-backbone architecture is updated by minimizing the following joint objective:
\begin{equation}
    \mathcal{L}_{\text{PPO}} = \mathcal{L}_{P}^{\text{CLIP}} + c_v \mathcal{L}_{V}^{\text{CLIP}} + \beta \mathbb{D}_{\text{KL}}(\pi_\theta \| \pi_{\text{ref}})
\end{equation}

\section{Experiment}

\begin{table*}[t]
\centering
\vspace{-0.5em}
\caption{Experimental results on R2R-CE and RxR-CE datasets.}
\label{tab:vln_ce_comparison}
\footnotesize
\setlength{\tabcolsep}{4pt}
\renewcommand{\arraystretch}{0.8}
\begin{tabular}{l c c c cccc cccc cccc}
\toprule
\multirow{2}{*}{Method} & \multirow{2}{*}{Params} & \multirow{2}{*}{Training} & \multirow{2}{*}{Action Space} & \multicolumn{4}{c}{Observation} & \multicolumn{4}{c}{R2R Val-Unseen} & \multicolumn{4}{c}{RxR Val-Unseen} \\
\cmidrule(lr){5-8} \cmidrule(lr){9-12} \cmidrule(lr){13-16}
 & & & & S.RGB & M.RGB & Depth & Odo. & NE$\downarrow$ & OS$\uparrow$ & SR$\uparrow$ & SPL$\uparrow$ & NE$\downarrow$ & SR$\uparrow$ & SPL$\uparrow$ & nDTW$\uparrow$ \\
\midrule
R2R-CMTP \cite{chen2021cmtp} & $\sim$0.3B & IL & Macro & & \checkmark & \checkmark & \checkmark & 7.90 & 38.0 & 26.4 & 22.7 & - & - & - & - \\
ETPNav \cite{an2024etpnav} & $\sim$0.3B & IL & Macro & & \checkmark & \checkmark & \checkmark & 4.71 & 65.0 & 57.0 & 49.0 & 5.64 & 54.8 & 44.9 & 61.9 \\
BEVBert \cite{an2022bevbert} & $\sim$0.3B & IL & Macro & & \checkmark & \checkmark & \checkmark & 4.57 & 67.0 & 59.0 & 50.0 & - & - & - & - \\
HNR \cite{hnr} & $\sim$0.3B & IL & Macro & & \checkmark & \checkmark & \checkmark & 4.42 & 67.0 & 61.0 & 51.0 & 5.51 & 56.4 & 46.7 & 63.6 \\
\midrule
VLN-R1 \cite{qi2025vlnr1} & 7B & IL+RL & Micro & \checkmark & & & & 7.00 & 41.2 & 30.2 & 21.8 & 10.4 & 22.3 & 17.5 & - \\
MapNav \cite{zhang2025mapnav} & 7B & IL & Micro & \checkmark & & & & 4.93 & 53.0 & 39.7 & 37.2 & 7.62 & 32.6 & 27.7 & 43.5 \\
UniNaVid \cite{zhang2024uninavid} & 7B & IL & Micro & \checkmark & & & & 5.58 & 53.3 & 47.0 & 42.7 & 6.24 & 48.7 & 40.9 & - \\
StreamVLN \cite{wei2025streamvln} & 7B & IL & Micro & \checkmark & & & & 4.98 & 64.2 & 56.9 & 51.9 & 6.22 & 52.9 & 46.0 & 61.9 \\
\midrule
Ours-DAgger & 0.5B & IL & Macro & & \checkmark & \checkmark & \checkmark & 4.06 & 70.1 & 65.1 & 53.5 & 5.20 & 59.3 & 48.9 & 63.9 \\
\textbf{Ours-RFT} & 0.5B & IL+RL & Macro & & \checkmark & \checkmark & \checkmark & \textbf{3.89} & \textbf{72.6} & \textbf{68.1} & \textbf{57.3} &\textbf{5.07} & \textbf{62.3} & \textbf{50.2} & \textbf{66.8} \\
\bottomrule
\end{tabular}
\vspace{-1em}
\end{table*}

\subsection{Experimental Setup}

\subsubsection{Datasets and Metrics}
We evaluate our approach on two standard benchmarks in the Habitat \cite{savva2019habitat} simulator: R2R-CE \cite{VLN} and RxR-CE \cite{RXR}. R2R-CE features concise English instructions guiding agents along shortest paths, while RxR-CE contains highly descriptive, multilingual instructions with longer, potentially non-shortest paths. We adopt a standard suite of navigation metrics \cite{anderson2018evaluation, ilharco2019general}: Navigation Error (NE), Success Rate (SR), Oracle SR (OSR), Success weighted by Path Length (SPL), normalized Dynamic Time Warping (nDTW).

\subsubsection{Implementation Details}
We employ the ViT-B/32 \cite{dosovitskiy2020image} pretrained on CLIP \cite{radford2021clip} as RGB encoder and the ResNet-50 \cite{he2016deep} pretrained on DD-PPO \cite{wijmans2019ddppo} as depth encoder, alongside the RoBERTa text encoder \cite{liu2019roberta}. The model features a 2-layer panorama encoder and a 4-layer Cross-Modal Fusion Network, with a unified hidden dimension of $d=768$. Across all training stages, the visual encoders are frozen, while all other modules remain trainable.

The pretraining stage leverages a unified dataset combining five sources: Prevalent \cite{hao2020towards}, Prevalent\_LVLM\_Aug, RxR-Marky\cite{wang2022marky}, and the original R2R\_train \cite{VLN} and RxR\_train \cite{RXR}. Specifically, Prevalent\_LVLM\_Aug is constructed by utilizing the LVLM API to re-annotate the trajectories in Prevalent, which enhances the linguistic diversity. Subsequent DAgger and RFT stages are conducted separately on the R2R-CE and RxR-CE datasets.

For the closed-loop RFT stage, We collect 256 trajectories before performing an update. Each update consists of 1 optimization epoch, running for a total of 500 iterations. Other standard PPO hyperparameters include a discount factor $\gamma=0.99$, a GAE parameter $\lambda=0.95$, a clipping range $\epsilon=0.2$, a value loss coefficient $c_v=0.5$, and a KL loss weight $\beta=0.04$. All training is conducted on 4 NVIDIA RTX 5880 GPUs. The Pretraining and DAgger stages require approximately 72 hours each, while the efficient RFT stage takes only about 16 hours. During inference, our high-level planner takes only 0.05s to make a macro decision.

\subsection{Comparison with State-of-the-Art Methods}
Table \ref{tab:vln_ce_comparison} presents the comparison of our approach against previous state-of-the-art methods on the R2R-CE and RxR-CE Val-Unseen splits.

While our pre-RFT model (Ours-DAgger) already establishes a remarkably strong initialization—surpassing both graph-based baselines (e.g., HNR) and 7B LVLM approaches (e.g., StreamVLN)—our primary focus is the effectiveness of the subsequent RL stage. Specifically, the RFT stage yields comprehensive and consistent improvements over the DAgger baseline. On the R2R-CE dataset, it boosts the SR from 65.1\% to 68.1\% and SPL from 53.5\% to 57.3\%. Similar gains are observed on the more complex RxR-CE dataset (e.g., SR reaches 62.3\% and nDTW reaches 66.8\%). These enhancements demonstrate that applying RL exclusively to the macro MDP effectively improves performance over IL models. In contrast, other methods under macro action space did not utilize closed-loop RFT to achieve further performance improvements \cite{chen2021cmtp, an2024etpnav, an2022bevbert, hnr}, while LVLM-based methods applied open-loop RFT on the micro action space but yielded unsatisfactory results \cite{qi2025vlnr1}.

\subsection{Ablation Studies on the RL Framework}
To validate our proposed RFT framework, we conduct ablation studies on the R2R-CE Val-Unseen split. Table \ref{tab:rl_ablation} summarizes the results and training cost, with $\Delta$ denoting absolute improvements over the DAgger baseline.

\begin{table}[htbp]
\centering
\caption{Ablation of RL algorithms, architectural designs, and internal settings. RFT time is measured in hours.}
\label{tab:rl_ablation}
\resizebox{\columnwidth}{!}{%
\begin{tabular}{l c c c c c}
\toprule
Method  & SR$\uparrow$ & SPL$\uparrow$ & Time(h)$\downarrow$ & $\Delta$ SR$\uparrow$ & $\Delta$ SPL$\uparrow$ \\
\midrule
Ours-DAgger & 65.1 & 53.5 & - & - & - \\
\midrule
\multicolumn{6}{l}{\textbf{Algorithm 1: PPO}} \\
~~ \textbf{Default (Ours-RFT)} & \textbf{68.1} & \textbf{57.3} & \textbf{$\sim$16} & \textbf{+3.0} & \textbf{+3.8} \\
~~ w/ History-only Value Critic & 66.2 & 54.7 & $\sim$25 & +1.1 & +1.2 \\
~~ w/ Dense Soft Reward & 67.4 & 56.5 & $\sim$16 & +2.3 & +3.0 \\
~~ w/ Multi-Epoch Update & 66.8 & 55.4 & $\sim$16 & +1.7 & +1.9 \\
\midrule
\multicolumn{6}{l}{\textbf{Algorithm 2: GRPO}} \\
~~ Default & 66.8 & 55.8 & $\sim$16 & +1.7 & +2.3 \\
~~ w/ Multi-Epoch Update & 65.9 & 54.8 & $\sim$16 & +0.8 & +1.3 \\
\midrule
\multicolumn{6}{l}{\textbf{Algorithm 3: SAC}} \\
~~ Default & 67.2 & 56.1 & $\sim$21 & +2.1 & +2.6 \\
~~ w/ Dense Soft Reward & 66.2 & 55.5 & $\sim$21 & +1.1 & +2.0 \\
\bottomrule
\end{tabular}%
}
\vspace{-10pt}
\end{table}

\textbf{The Action-Aware Value Head.} We first ablate our value head design by substituting it with a \textit{History-only Value Critic} that strictly observes the explored trajectory $\mathcal{H}_t$ without seeing the generated frontiers. Because its input structure diverges from the policy head, it cannot share the cross-modal backbone, necessitating a redundant, detached value network. As shown in Table \ref{tab:rl_ablation}, this topological blindness yields a dual penalty: performance gains shrink (+1.1 $\Delta$SR vs. +3.0 $\Delta$SR), and training time inflates (from 16 to 25 hours). This confirms that explicitly observing the macro action space is essential for the critic to accurately penalize dead-end topological states without incurring unnecessary computational overhead.

\textbf{RL Algorithm Comparison.} Keeping the total trajectory samples consistent, we compare PPO against GRPO and a discrete variant of Soft Actor-Critic (SAC). GRPO (group size $G=8$) offers a straightforward optimization but lacks a critic, depriving it of fine-grained, step-level credit assignment, resulting in moderate gains (+1.7 $\Delta$SR). SAC outperforms GRPO (+2.1 $\Delta$SR) because its Q-networks score specific frontier tokens, enabling step-level evaluation. However, it falls short of PPO mainly due to two flaws. First, its off-policy experience replay injects stale representations from outdated policies, destabilizing the cross-modal alignment. Second, while SAC's online Q-networks can share the policy's backbone, computing stable TD targets requires maintaining a delayed copy of this multimodal backbone, incurring a computational penalty ($\sim$21h).

\textbf{Reward Formulation and Update Epochs.} Finally, we ablate two internal configurations across different algorithms. Regarding the reward, substituting our outcome-driven sparse reward with a Dense Soft Reward (reduction in geodesic distance) yields sub-optimal performance for both PPO and SAC. While dense rewards intuitively guide the agent toward the destination, they fail to explicitly enforce instruction fidelity. Consequently, the agent tends to exploit greedy geometric shortcuts rather than strictly following the language commands. Regarding update frequency, Strict On-Policy Updates consistently outperform multi-epoch updates (Epochs=2) on the same sampled batch for both PPO and GRPO. In the RFT stage, performing multiple gradient steps on a limited batch of trajectories leads to representation overfitting. Even with PPO's clipping mechanisms, aggressively reusing the same batch over-optimizes the cross-modal backbone to local environmental noise, ultimately degrading the generalizable visual-linguistic priors.

\subsection{Comparison with Micro Action Space Paradigms}
To validate that macro action spaces are a crucial prerequisite for effective RL in VLN-CE, we construct a baseline for micro action space, using an identical visual-linguistic backbone. 
These two baselines share the same training process up to the DAgger stage, after which they diverge into distinct RFT paradigms operating at the motor action granularity: open-loop and closed-loop.

\textbf{Baseline Setup.} To ensure a fair comparison, we adapt our topological representation for low-level control while preserving the identical cross-modal backbone. Frontier nodes are retained as spatial features rather than executable actions, and a new explored node is dynamically added only when the agent moves 3 meters away from existing nodes. The policy head is replaced with a GRU that autoregressively predicts the next four motor actions. We pretrain and DAgger-tune this architecture using a Time-Decayed cross-entropy loss over the 4-step sequence (assigning higher weights to immediate actions, similar to \cite{qi2025vlnr1}). Both the open-loop and closed-loop RFT variants are initialized from this same DAgger checkpoint, and we ensure the total number of sampled trajectories matches our high-level PPO training.

\begin{table}[t]
\centering
\caption{Results of RFT under micro action space.}
\label{tab:low_level_comparison}
\resizebox{\columnwidth}{!}{%
\begin{tabular}{l c c c c c}
\toprule
Paradigm & SR$\uparrow$ & SPL$\uparrow$ & Time(h)$\downarrow$ & $\Delta$ SR$\uparrow$ & $\Delta$ SPL$\uparrow$ \\
\midrule
\multicolumn{6}{l}{\textbf{Micro Action Space}} \\
~~ Pre-RFT (DAgger Baseline) & 31.1 & 26.4 & - & - & -\\
~~ Open-Loop RFT & 31.1 & 26.4 & $\sim$6 & +0.0 & +0.0\\
~~ Closed-Loop RFT & 31.7 & 26.6 & $\sim$120 & +0.6 & +0.2\\
\bottomrule
\end{tabular}%
}
\vspace{-10pt}
\end{table}

\begin{figure}[t]
    \centering
    \begin{minipage}[t]{1\linewidth}
    \centering
    \includegraphics[width=1\linewidth]{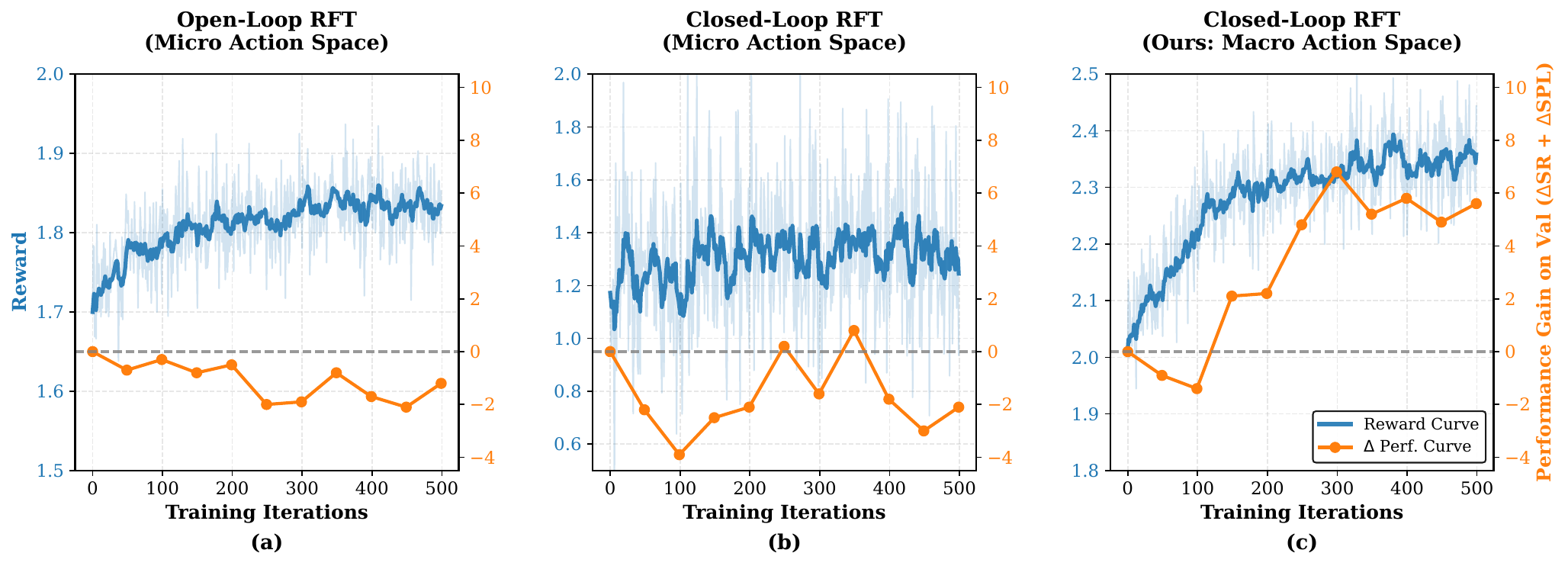}
    \end{minipage}
    \caption{\small{Training curves under three different RFT paradigms.}}
    \label{fig:training_curves}
    \vspace{-20pt}
\end{figure}


\textbf{Open-Loop RFT Performance.} Here the model is trained offline on fixed datasets with a Time-Decayed Reward \cite{qi2025vlnr1}, optimizing sequence generation without environmental feedback. As shown in Table \ref{tab:low_level_comparison} and Figure \ref{fig:training_curves}, this yields zero gain over the DAgger baseline: the reward rises while validation performance declines, indicating overfitting rather than generalization. Although prior work \cite{qi2025vlnr1} reports gains from similar techniques, we attribute these to compensating for flaws in its IL stage—namely the absence of temporal loss decay and teacher-forced token generation. Since our DAgger baseline already resolves both via autoregressive rollouts and a Time-Decayed loss, open-loop RFT over static datasets offers no further navigational advantage.

\textbf{Closed-Loop RFT Performance.} Here we execute the first predicted action in the environment and optimize it via PPO (using the identical reward in Equation \ref{eq:reward}), while the remaining three steps receive auxiliary DAgger supervision. Despite a 120-hour training period, this paradigm yields only marginal gains (+0.6 SR) and suffers from severe reward variance (Figure \ref{fig:training_curves}). The root cause is that treating every motor command as an independent RL step drastically extends the decision horizon, exacerbating credit assignment and rendering the sparse terminal reward uninformative for early decisions. Thus, while closed-loop feedback can correct minor policy drift, the excessively long action sequences prevent convergence to a robust policy.

\textbf{The Advantage of Macro Action Space.} In contrast, our topological formulation elevates the RL action space to macro waypoint selections, compressing the decision horizon from hundreds of micro steps to merely 5-20 choices. This structural abstraction naturally circumvents the credit assignment bottleneck, enabling our closed-loop RFT to converge and achieve substantial gains (Table \ref{tab:rl_ablation} and Figure \ref{fig:training_curves}). Ultimately, both the absolute performance and the relative improvements validate that macro action spaces are fundamental to unlocking the sample efficiency and effectiveness of closed-loop RL in VLN-CE.

\subsection{Orthogonality of RFT and Dataset Scaling for Performance Gains}
\begin{figure}[t]
    \centering
    \begin{minipage}[t]{1\linewidth}
    \centering
    \includegraphics[width=1\linewidth]{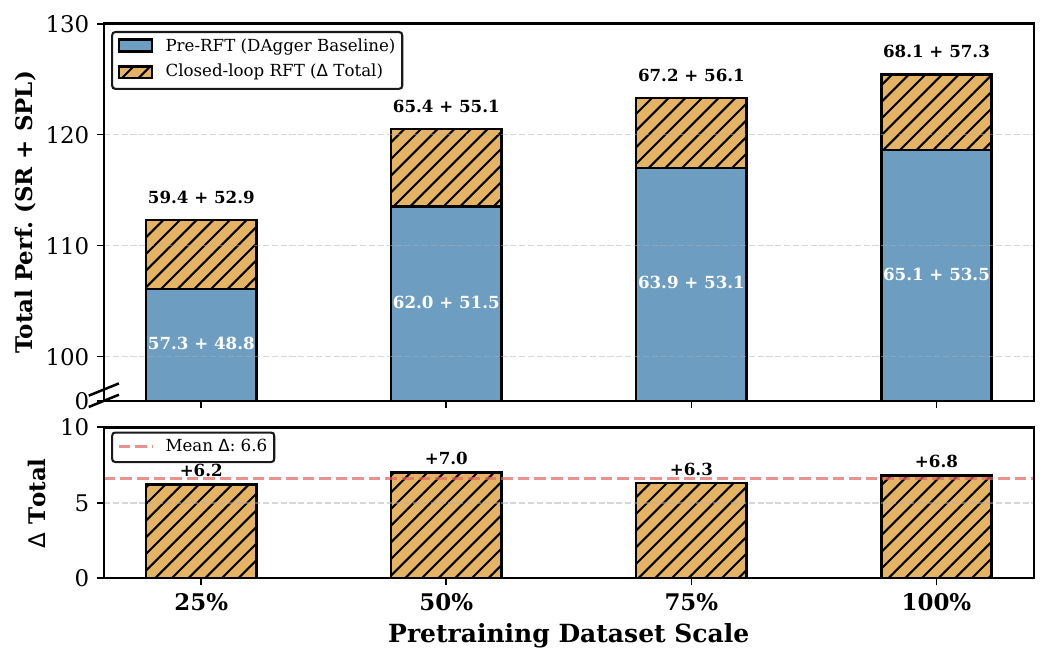}
    \end{minipage}
    \caption{\small{Performance Gains of RFT under different pretraining dataset scales.}}
    \label{fig:orthogonality}
    \vspace{-20pt}
\end{figure}

To further understand the role of our RFT stage, we investigate its relationship with the scale of the pretraining dataset. We train four base models using 25\%, 50\%, 75\%, and 100\% of the full pretraining dataset, respectively. Subsequently, each base model undergoes the DAgger and RFT pipeline. Figure \ref{fig:orthogonality} illustrates the total navigation performance (defined as the sum of SR and SPL) for both the DAgger baselines and their corresponding RFT-enhanced versions.

As expected, scaling the pretraining dataset yields substantial improvements in the base DAgger models. However, the most critical observation is the behavior of the RFT stage: regardless of the base model's initial competency, RFT delivers a consistent performance boost. The absolute performance gain ($\Delta$ Total) remains stable across all scales, hovering tightly between +6.2 and +7.0.

This consistent delta proves a fact: the performance gains from RFT and dataset scaling are orthogonal. It confirms that RFT doesn't act as a "capacity patch" to compensate for weak base models' lack of navigational knowledge. Instead, it addresses the inherent mismatch between DAgger's supervision signals and instruction fidelity—a structural flaw rooted in IL paradigm that scaling datasets cannot resolve.

\begin{figure*}[t]
    \centering
    \includegraphics[width=\textwidth]{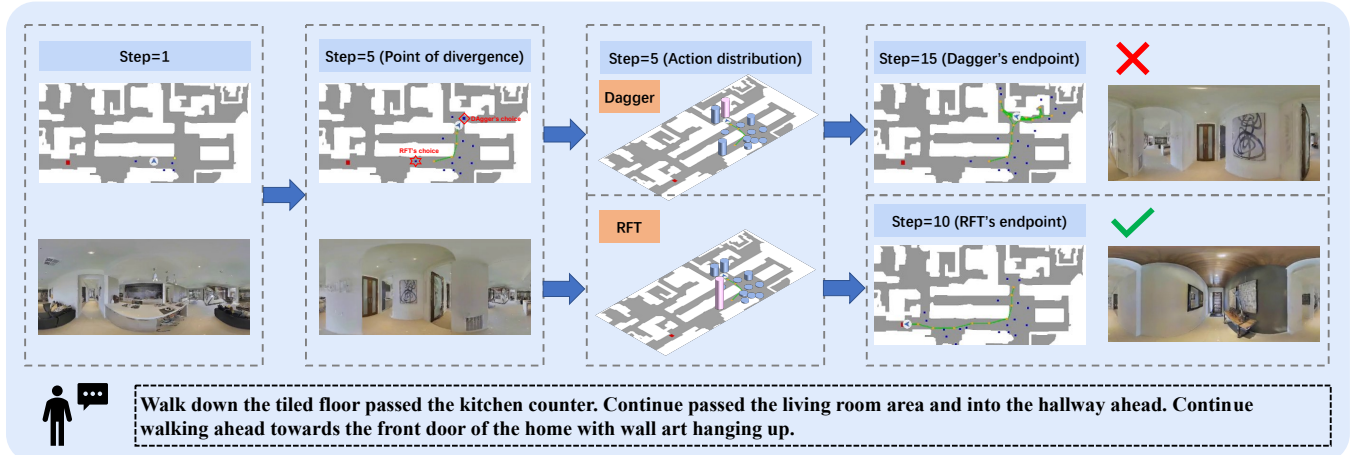}
    \vspace{-15pt}
    \caption{A representative episode to show the differences in policy and action distribution between DAgger and RFT models.}
    \label{fig:case study}
    \vspace{-10pt}
\end{figure*}

\subsection{Case Study}
Figure \ref{fig:case study} illustrates the policy differences between the RFT and the DAgger model through a representative episode. Due to initial heading ambiguity, both models move in the wrong direction. The divergence occurs at step 5. The RFT model recognizes the trajectory-instruction mismatch (the door in step 5 is not like the front door) and leverages the topological memory to select a frontier node generated at starting position. It successfully backtracks and completes the task. In contrast, the DAgger model falls into a wandering pattern and ultimately fails.

This behavioral difference is explicitly reflected in the action distributions at the divergence step. The DAgger model exhibits a flat distribution, indicating it senses a mismatch but cannot confidently identify the corrective action. Conversely, the RFT model assigns a dominant probability to the correct backtrack node while suppressing the incorrect node's probability. This confirms that RFT effectively re-ranks the graph's top candidate actions based on reward signals, achieving better trajectory-instruction alignment.

\section{Conclusion}
In this work, we reformulate the VLN-CE task as a hierarchical decision-making problem modeled by a Hierarchical MDP, which breaks through the sample-efficiency bottleneck of applying closed-loop RL. By abstracting the continuous environment into a topological graph and elevating the action space to high-level frontier selection, we reduce the decision horizon and make RL optimization tractable. Through our closed-loop Reinforcement Fine-Tuning (RFT) paradigm via PPO, the agent improves its long-horizon instruction-following capabilities, achieving state-of-the-art performance on the R2R-CE and RxR-CE datasets.

While effective, our framework relies on a waypoint predictor, which inherently restricts the macro action space to its output distribution rather than guaranteeing theoretical global reachability. Future work will explore globally reachable macro action spaces and bridge the sim-to-real gap by replacing the heuristic low-level controller with robust local navigation algorithms for physical robot deployment.



\bibliography{root}

\bibliographystyle{ieeetr}

\end{document}